\documentclass[sigconf]{acmart}

\AtBeginDocument{%
  \providecommand\BibTeX{{%
    \normalfont B\kern-0.5em{\scshape i\kern-0.25em b}\kern-0.8em\TeX}}}

\setcopyright{acmcopyright}
\copyrightyear{2018}
\acmYear{2018}
\acmDOI{XXXXXXX.XXXXXXX}

\acmConference[Conference acronym 'XX]{Make sure to enter the correct
  conference title from your rights confirmation emai}{June 03--05,
  2018}{Woodstock, NY}
\acmPrice{15.00}
\acmISBN{978-1-4503-XXXX-X/18/06}

\begin{document}

\title{GreenDB: Toward a Product-by-Product Sustainability Database}

\author{Sebastian~Jäger}
\email{sebastian.jaeger@bht-berlin.de}
\affiliation{%
  \institution{Berliner Hochschule für Technik}
  \streetaddress{Luxemburger Str. 10}
  \city{Berlin}
  \country{Germany}
  \postcode{13353}
}

\author{Jessica~Greene}
\author{Max~Jakob}
\author{Ruben~Korenke}
\affiliation{%
	\institution{Ecosia GmbH}
	\city{Berlin}
	\country{Germany}
}

\author{Tilman~Santarius}
\affiliation{%
	\institution{Technical University of Berlin}
	\institution{Einstein Center Digital Future}
	\city{Berlin}
	\country{Germany}
}

\author{Felix~Bießmann}
\affiliation{%
	\institution{Berliner Hochschule für Technik}
	\institution{Einstein Center Digital Future}
	\city{Berlin}
	\country{Germany}
}

\newcommand{\code}[1]{\texttt{#1}}
\newcommand{\felix}[1]{\textcolor{red}{[Felix: #1]}}
\newcommand{\sebastian}[1]{\textcolor{blue}{[Sebastian: #1]}}

%\affiliation{%
%  \institution{Inria Paris-Rocquencourt}
%\affiliation{%
% \institution{Rajiv Gandhi University}
%\affiliation{%
%  \institution{Tsinghua University}
%  \streetaddress{30 Shuangqing Rd}
%\affiliation{%
%  \institution{Palmer Research Laboratories}
%  \streetaddress{8600 Datapoint Drive}
%  \country{USA}
%\affiliation{%
%  \institution{The Th{\o}rv{\"a}ld Group}
%  \streetaddress{1 Th{\o}rv{\"a}ld Circle}
%\email{jsmith@affiliation.org}
%\affiliation{%
%  \institution{The Kumquat Consortium}
%  \city{New York}
%\email{jpkumquat@consortium.net}


\begin{abstract}
	The production, shipping, usage, and disposal of consumer goods have a substantial impact on greenhouse gas emissions and the depletion of resources. 
	Modern retail platforms rely heavily on Machine Learning (ML) for their search and recommender systems. Thus, ML can potentially support efforts towards more sustainable consumption patterns, for example, by accounting for sustainability aspects in product search or recommendations. However, leveraging ML potential for reaching sustainability goals requires data on sustainability. Unfortunately, no open and publicly available database integrates sustainability information on a product-by-product basis. In this work, we present the \emph{GreenDB}, which fills this gap. Based on search logs of millions of users, we prioritize which products users care about most. The GreenDB schema extends the well-known \emph{schema.org Product} definition and can be readily integrated into existing product catalogs to improve sustainability information available for search and recommendation experiences. We present our proof of concept implementation of a scraping system that creates the GreenDB dataset.
\end{abstract}

% TODO

% TODO: ok?

\maketitle

\section{Introduction}

Climate change is one of the most important challenges of our generation. For the average citizen in western civilizations, fighting climate change requires, next to many other efforts, primarily one thing: People have to change their consumption behavior! The increasing demand for consumer goods and their implications for globalized supply chains, shipping, and disposal are major factors in increased greenhouse gas emissions \cite{CO2_Emission_Trend, BMU_co2} and the depletion of resources. With every purchase decision made, people can decide for -- or against -- more sustainable consumption behavior.
While many consumers would like to choose sustainable options \cite{PWCSurvey, WWF2021, BMU_umweltbewustsein} at the point of sale, the relevant information is not available. This is at odds with research efforts invested by major retailers into improving the data quality of their product catalogs as well as search and recommendation experiences. Leveraging Machine Learning (ML) for enriching product catalogs is the focus of scientists at these companies \cite{Biessmann2018, Biessmann2019}, and there are attempts to take into account sustainability information and user preferences into recommender systems \cite{Tomkins20818}. Our research and discussions with researchers, managers, and engineers of the largest German online retailers have demonstrated that the most important factor hindering the integration of sustainability aspects in their shopping sites indeed is the availability of trustworthy sustainability data. In this study, we set out to tackle this challenge and present a first release of what we refer to as \emph{GreenDB}.
%For this reason, we propose to build a \emph{GreenDB}, a database with transparent sustainability information about products.

%\felix{The rise of ML technology in the past decades has clearly shown that without (high quality) data} it is a fundamental prerequisite for building robust machine learning (ML) systems. A commonly used example is the ImageNet dataset, a large-scale open and publicly available image dataset \cite{ImageNet}. Its availability boosted the research in computer vision (CV) and increased the CV model performance heavily, also known as the "ImageNet moment". Unfortunately, not all ML sub-fields can profit from their own "ImageNet moment".
%One of the largest societal problem at the moment is climate change. While many consumers would like to choose more sustainable options \cite{PWCSurvey, WWF2021, BMU_umweltbewustsein} at the point of sale to decrease their footprint, the relevant information is not available.
%For this reason, we build the \emph{GreenDB}, a database with transparent sustainability information about products. 

This work is not the first open data initiative of sustainable information. Some sustainability labels provide lists of certified products publicly\footnote{
	For example,  \href{https://ec.europa.eu/info/energy-climate-change-environment/standards-tools-and-labels/products-labelling-rules-and-requirements/energy-label-and-ecodesign/product-database_en}{European Product Database for Energy Labelling (\emph{EPREL})} or \href{https://www.blauer-engel.de/}{\emph{Blue Angle}}
}. However, these mostly contain product categories that do not play a role in consumers' everyday lives and are often not machine-readable.
Some large retailers present products' sustainable information for marketing purposes. However, those companies have commercial interests, which raises concerns about the data's trustworthiness. To summarize, we could not find a trustworthy and comprehensive resource of sustainability information on a product-by-product basis.

In this paper, we propose a schema, which extends the widely adopted schema.org definition, and a proof of concept implementation of a scraping system that creates the GreenDB. To infer sustainability information for products, we rely on sustainability labels that certify these. In the end, we show details about the first public available GreenDB dataset.

%\begin{enumerate}
%	\item Data is important for ML
%	\item E.g., ImageNet boosted research in CV field, known as ImageNet moment
%	\item More sustainable consumption is important to fight climate change
%	\item There are just a few product datasets, more in Section \ref{sec:related_work}
%	\item However, no dataset that has sustainability information of products
%	\item Therefore, we propose a) a schema for such Sustainability Database (GreenDB)
%	\item b) a system to create such a dataset.
%	\item c) publish the first openly available sustainable products dataset
%\end{enumerate}
%
%Here, we should define sustainability labels (reference)
\section{Related Work}
\label{sec:related_work}

Some available product datasets are more or less similar to what we do:
\begin{enumerate}
	\item \emph{Products-10K - Large Scale Product Recognition Dataset} \cite{Products10k}
	\item \emph{Amazon Review Data (2018)} \cite{AmazonReviewData}
	\item \emph{Flipkart Products} competition on Kaggle\footnote{\url{https://www.kaggle.com/PromptCloudHQ/flipkart-products}}
	\item \emph{Otto Group Product Classification Challenge} competition on Kaggle\footnote{\url{https://www.kaggle.com/c/otto-group-product-classification-challenge/}}
\end{enumerate}

Since it is designed for an image recognition task, the Products-10K dataset differs most from the GreenDB. It consists of roughly $150,000$ images of about $10,000$ different products and their categories. The goal is to predict products' categories based on their images. \cite{Products10k}

The dataset published for the Otto Group Product Classification Challenge is tabular. It has $93$ features for more than $200,000$ products. The target class is similar to Products-10K to distinguish the products between categories.

As its name suggests, the Amazon Review Data (2018) focuses on products reviews. However, it also contains product attributes such as \emph{color} or \emph{size} and offers many product categories. Since the dataset spans from 1996 to 2018, the amount of data is enormous (about $34GB$). \cite{AmazonReviewData}

Flipkart Products is most similar to GreenDB. It is a tabular dataset with $20,000$ products and 15 columns. These attributes are partially very similar to ours, e.g., \emph{product\_name}, \emph{product\_url}, \emph{retail\_price}, \emph{description}, and \emph{brand}.

However, all mentioned datasets have in common that they do not integrate products' sustainability information, which is our main focus.

\section{Methodology}
\label{sec:methodology}

To create the GreenDB, we need to solve two main problems. Firstly, the collection of products and their attributes, and secondly, we need sustainability information. For this, we leverage that some large online retailers present products' sustainability information, e.g., materials the product is made of, the energy consumption of electronic devices, and, most importantly, sustainability labels that certify the product. We exploit that the number of sustainability labels is much smaller than the number of available products. This allows us to infer sustainability information for a large number of products by evaluating a relatively small number of sustainability labels.

\paragraph{Sustainability label evaluation} The International Organization for Standardization (ISO) 14020 series defines principles and three types for voluntary sustainability labels, two of which are relevant for this work. Type I requires a third-party verification (ISO 2018) \cite{Typ_I_sustainability_labels}, and Type II are self-declarations without independent verification (ISO 2016) \cite{Typ_II_sustainability_labels}. In the following, we refer to them as sustainability labels if their type does not matter, \emph{third-party labels} for Type I, and \emph{private labels} for Type II.\\
We use the \emph{Sustainability Standards Comparison Tool} (SSCT) for evaluating third-party labels. Its development was initially initiated by the German government.
SSCT evaluates both the (third-party) labels' aspects and their implementation systems, leading to three scores ($0 - 100$) that represent their evaluation in the dimensions of \emph{credibility}, \emph{environment}, and \emph{socio-economic}. \cite{Siegelklarheit_Tool}

\subsection{GreenDB schema}
\label{sec:schema}

% Use a single table environment to keep all tables at one place
\begin{table*}[h]
	\begin{tabular}{lllllll}
		\toprule
		\textbf{GTIN} &
		\textbf{Name} &
		\textbf{Description} &
		\textbf{Manufacturer} &
		\textbf{Category} &
		\textbf{Sustainability-Labels} \\
		\midrule
		4250805445834&RAIKOU Sweatjacke (...)&Sweatjacke Unisex (...)&RAIKOU&JACKET&[MIG\_OEKO\_TEX]\\
		\bottomrule
	\end{tabular}
	\caption{\textit{Products}. Representations of products that follow Schema.org's Product definition. The attribute \emph{Sustainability-Labels} is an extension and models that the product is certified by a set of \emph{Sustainability Labels} (or does not have certificates).}
	\label{tab:greenDB:products}
	
	\begin{tabular}{lllrrr}
		\toprule
		\textbf{Sustainability-LabelID} &
		\textbf{Name} &
		\textbf{Description} &
		\textbf{Cred.} &
		\textbf{Env.} &
		\textbf{Socio-Eco.} \\
		
		\midrule
		MIG\_OEKO\_TEX&OEKO-TEX Made in Green&Nachverfolgbares Produktsiegel (...)&76&80&80\\
		\bottomrule
	\end{tabular}
	\caption{\textit{Sustainability Labels}. Representations of sustainability labels and their SSCT evaluation scores as presented on Siegelklarheit. Abbreviations: \emph{Cred.} stands for Credibility, \emph{Env.} for Environment, and \emph{Socio-Eco.} for Socio-Economic.}
	\label{tab:greenDB:labels}
\end{table*}
\label{tab:greenDB}

The GreenDB schema is highly inspired by \emph{Schema.org's Product}\footnote{\url{https://schema.org/Product}} definition. Schema.org is a community activity to create definitions for structured data on the Internet.
The table \emph{Products}, shown in Table \ref{tab:greenDB:products}, extends this and introduces the attribute \emph{Sustainability-Labels}, a set of foreign keys referencing the table \emph{Sustainability Labels}, shown in Table \ref{tab:greenDB:labels}.
This reference allows inferring sustainability information for the given product.
However, all other properties from Schema.org Product are still valid.
Table \emph{Sustainability Labels} (see Table \ref{tab:greenDB:labels}) represents sustainability labels and their SSCT evaluation scores.
As a starting point, we use the website \emph{Siegelklarheit}\footnote{\url{https://www.siegelklarheit.de}}, which applies SSCT to third-party labels and present their results publicly.

\subsection{Scraping system}
\label{sec:scraping_system}

\begin{figure}[h]
	\centering
	\includegraphics[width=\linewidth]{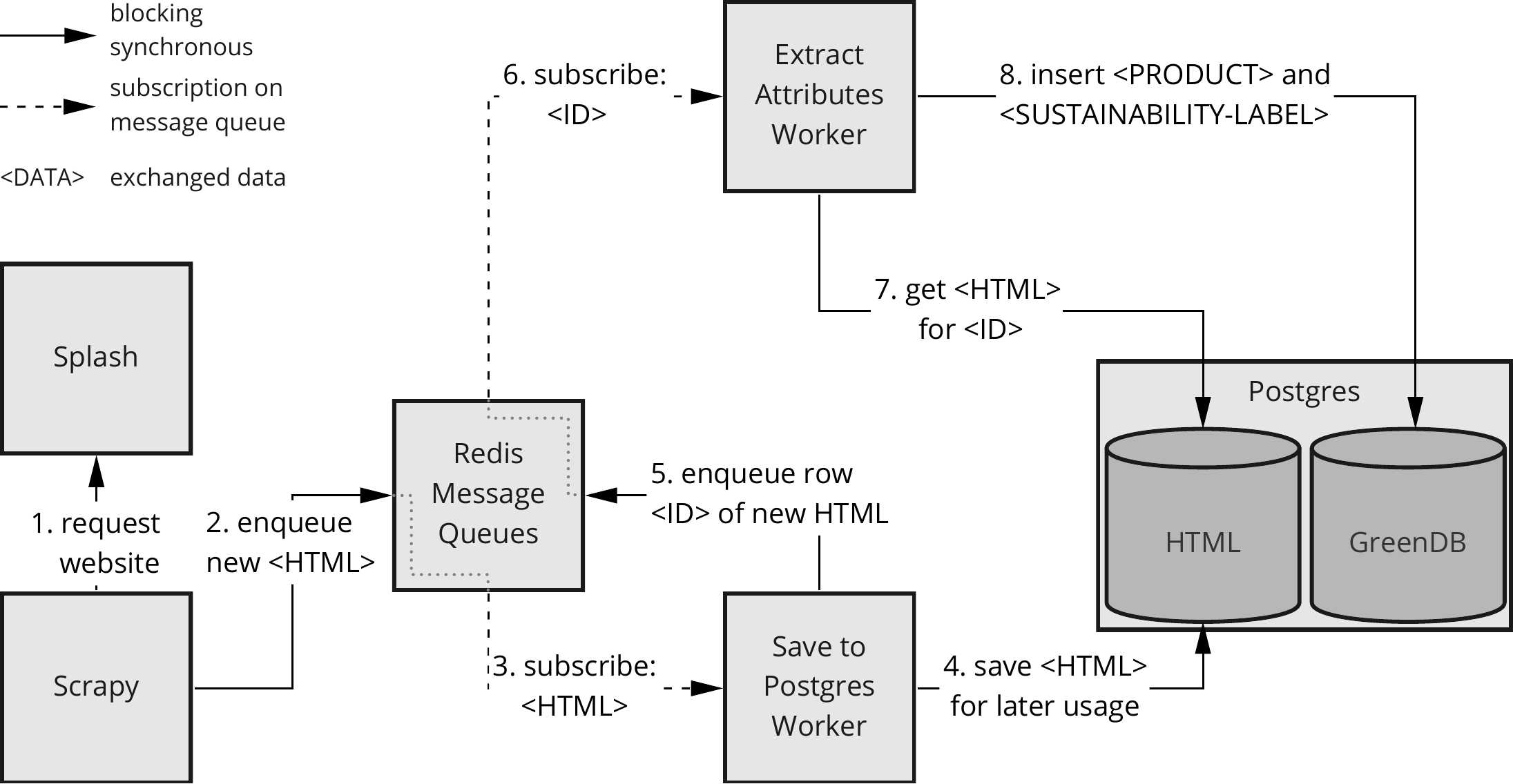}
	\caption{GreenDB's scraping system. Sketch of the main components and their communication paths.}
	\label{fig:scraping_system}
\end{figure}

To implement a robust and extensible scraping system that can be used to regularly get new products or versions of existing products, we use the framework \emph{scrapy}\footnote{\url{https://scrapy.org}}. It respects the websites' \code{robot.txt} rules and rate limits. Since many modern websites use JavaScript to load their content asynchronously while scrolling through their page, we use \emph{splash}\footnote{\url{https://splash.readthedocs.io/en/stable/}} to render the JavaScript and, if necessary, interact with the websites.

After downloading the websites' HTML, we save them into a database. This reduces the number of requests to the web servers since we cache the necessary (unstructured) information and can re-use them at any time.
For increasing the scraping system's resilience and internal communications, we use \emph{RQ}\footnote{\url{https://python-rq.org/}} (Redis Queue). RQ makes it easy to use non-blocking and asynchronous message queues. The sending component (e.g., step two or four in Figure \ref{fig:scraping_system}) adds (unstructured) information to a specified queue, and, on the other hand, \emph{workers} can subscribe to these queues (e.g., step three or six in Figure \ref{fig:scraping_system}), receive the information and handle it. Since RQ executes the workers in fail-safe environments, errors do not break the scraping system, and the information is not lost.

Currently, there are two workers: a) a worker that simply caches the HTML files in a \emph{PostgreSQL}\footnote{\url{https://www.postgresql.org}}  database and b) a worker responsible for extracting product attributes and sustainability information from those HTML files and, finally, creates new entries in the GreenDB.
An increasing number of retailers are supporting schema.org attributes \cite{WDC}. We use the library \code{extruct}\footnote{\url{https://github.com/scrapinghub/extruct}} to extract general product information. Since sustainability information is not part of schema.org’s Product definition yet,  we use \emph{BeautifulSoup}\footnote{\url{https://www.crummy.com/software/BeautifulSoup/bs4/doc/}} to parse the websites' HTML and extract sustainability labels.

The code of our scraping system is available on GitHub: \url{https://github.com/calgo-lab/green-db}.

\section{GreenDB}

The first public release of our GreenDB dataset \cite{Jaeger_Biessmann_GreenDB_Data} is available on Zenodo: \url{https://zenodo.org/record/6078039}.

\paragraph{Fashion and electronics are the most important categories} We use hundreds of millions of German product search queries from a large search engine (about $6\%$ of all search queries). We manually labeled the $200$ most frequent search queries and found \emph{fashion} ($19\%$) and \emph{electronics} ($17\%$) are the most important categories. This roughly reflects their revenue on the German online market in recent years \cite{Warengruppen}.

We test our proof of concept implementation (see Section \ref{sec:scraping_system}) based on these two product categories and two of the top three largest German online stores: \emph{OTTO}\footnote{\url{https://www.otto.de}} and \emph{Zalando}\footnote{\url{https://www.zalando.de/}}\cite{German_Online_Shops_Size}. Both present products' sustainability information for a wide range of products. Zalando advertises about $48.000$ fashion products and Otto about $15.000$ fashion and $1.000$ electronic products as sustainable.

\subsection{Data scope}

Between February 17th and March 6th, we started our scraping system (described in Section \ref{sec:scraping_system}) six times. Multiple runs help to increase the number of products eventually integrated into the GreenDB because it is possible that errors, e.g., connection problems, blocked requests, or malformed HTML, lead to invalid products and, therefore, they are not available in the GreenDB. Based on manually examining both websites, we came up with $26$ product categories ($18$ fashion and $8$ electronics categories). For better accessibility for other researchers, we plan to use an open product taxonomy in future releases. 

To bootstrap the sustainability labels table, we used all third-party labels available on Siegelklarheit. We iteratively added all found but not yet integrated third-party labels into the database. This, finally, led to $142$ third-party labels, where $34$ of them are already evaluated by Siegelklarheit.

\subsection{Data volume and distribution}

The GreenDB dataset \cite{Jaeger_Biessmann_GreenDB_Data} consists of more than $17,000$ unique products. If a product is assigned to different categories on the online store, it is represented as multiple rows. For this reason, the total number of rows in the \code{products} file is larger, i.e., more than $21,000$. In the following, percentage values refer to the number of unique products if not stated otherwise. The products are roughly equally distributed from Zalando ($\sim 48\%$) and Otto ($\sim 52\%$). Since Zalando specializes in fashion products, all electronics products ($277$) are from OTTO.

Interestingly, the total number of as sustainable advertised products on the online store's website strongly differs from the actual scraped products. One reason for this is the partially very fine-grained category filters. We, at first, focused on the largest product categories and did not integrate niche products.
\begin{figure}[h]
	\centering
	\includegraphics[width=0.8\linewidth]{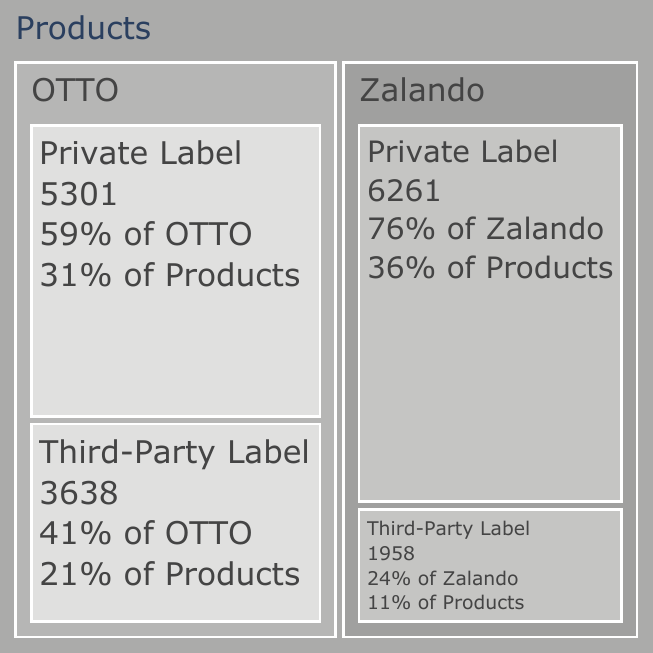}
	\caption{GreenDB dataset distribution. The nested rectangles represent the fractions of products depending on the attributes \emph{merchant} and \emph{sustainability labels}.  Sustainability labels are grouped depending on whether they are third-party or private labels.}
	\label{fig:greenDB_distribution}
\end{figure}

Surprisingly, more than two-thirds ($\sim 67\%$) of the products have private labels, e.g., "made with recycled materials" or "energy-saving device". In detail, these are $\sim 59\%$ of OTTO's and $\sim 76\%$ of Zalando's products.
Figure \ref{fig:greenDB_distribution} visually presents the data's nested distribution.
For private labels, we use \code{OTHER} as a representation of products advertised as sustainable. For transparency and to enable researchers to handle those products by themselves, e.g., by evaluating that information separately, we kept these products in the GreenDB dataset.

\section{Conclusion and future work}

In this paper, we present the GreenDB, a sustainability database on a product-by-product basis. Firstly, we discussed our approach, which uses sustainability labels to infer sustainability information for products. The schema we propose extends the product definition from schema.org and allows us to model that sustainability labels can certify products. We then discussed our proof of concept implementation of a scraping system, which finds products  and fills the GreenDB. Lastly, we present the first publicly available release of the GreenDB dataset.

In future work, we plan to integrate other product categories and their sustainability labels. Further, we plan to integrate non-German markets, and as mentioned above, we are investigating the usage of an open product taxonomy to increase the accessibility for other researchers.

%\section{Acknowledgments}
%
%Identification of funding sources and other support, and thanks to
%individuals and groups that assisted in the research and the
%preparation of the work should be included in an acknowledgment
%section, which is placed just before the reference section in your
%document.
%
%This section has a special environment:
%\begin{verbatim}
%  \begin{acks}
%  ...
%  \end{acks}
%\end{verbatim}
%so that the information contained therein can be more easily collected
%during the article metadata extraction phase, and to ensure
%consistency in the spelling of the section heading.
%
%Authors should not prepare this section as a numbered or unnumbered {\verb|\section|}; please use the ``{\verb|acks|}'' environment.

%\section{Appendices}
%
%If your work needs an appendix, add it before the
%``\verb|\end{document}|'' command at the conclusion of your source
%document.
%
%Start the appendix with the ``\verb|appendix|'' command:
%\begin{verbatim}
%  \appendix
%\end{verbatim}
%and note that in the appendix, sections are lettered, not
%numbered. This document has two appendices, demonstrating the section
%and subsection identification method.

%\begin{acks}
%TODO
%\end{acks}

\bibliographystyle{ACM-Reference-Format}
\bibliography{green-db}
%%

%\appendix

\end{document}